\documentclass[11pt]{article}

\usepackage{acl2023}

\usepackage{microtype}
\usepackage{inconsolata}
\usepackage{graphicx}
\usepackage{url}
\usepackage{todonotes}
\usepackage{enumitem}
\usepackage{amsmath}
\usepackage{xcolor}
\usepackage{hyperref}
\usepackage{xurl}
\usepackage{booktabs}

\usepackage{fancyhdr}
\usepackage{fontspec} 
\usepackage{polyglossia}

\setmainlanguage{english}
\setotherlanguage{hindi}

\newfontfamily\devanagarifont[
  Script = Devanagari,
  Path = ./,
  Extension = .ttf,
  UprightFont = *-Regular,
  BoldFont = *-Bold
]{NotoSansDevanagari}

\title{From Lexicon to AI: A Structured-Data Pipeline for Specialized Conversational Systems in Low-Resource Languages}
\date{}
\author{Siddhant Hitesh Mantri \\
  NMIMS, Mumbai \\
  \texttt{siddhant.mantri22@nmims.in} \\\And
  Dhara Gorasiya \\
  CFILT, IIT Bombay \\
  \texttt{dharagorasiya@gmail.com} \\\AND  % <--- CHANGE THIS FROM \And TO \AND
  Malhar Kulkarni \\
  CFILT, IIT Bombay \\
  \texttt{malhar@hss.iitb.ac.in} \\\And
  Pushpak Bhattacharya \\
  CFILT, IIT Bombay \\
  \texttt{pb@cse.iitb.ac.in} \\}

\usepackage{booktabs}

\usepackage[most]{tcolorbox}

\newtcolorbox{promptbox}[1][]{
  colback=white,        % White background
  colframe=black,       % Black frame
  boxrule=0.5pt,        % Thin line like listings
  arc=0pt,              % Sharp corners
  fontupper=\small,     % Smaller font size as requested
  title={#1},           % Optional title
  coltitle=black,       % Black title text
  attach boxed title to top left={yshift=-2mm, xshift=2mm},
  boxed title style={colback=white, frame hidden},
  enhanced,
  breakable,            % Allows the box to split across pages
  top=10pt              % Space at the top
}

\begin{document}
\maketitle
\begin{abstract}
Low-resource languages face a critical challenge in AI development: creating specialized conversational systems without access to massive training corpora. We present a systematic methodology for transforming structured linguistic resources into specialized AI systems, demonstrating that expert-curated lexical databases can serve as effective foundations for conversational AI development. Our approach converts Hindi WordNet into 1.25 million diverse instruction-response pairs, fine-tunes a 12B-parameter language model using resource-efficient LoRA with 4-bit quantization. Evaluation through a Hindi language learning chatbot suggests that structured-knowledge-based systems can improve pedagogical effectiveness in educational conversational settings (91.0 vs. 79.4-83.6 for general-purpose models) while maintaining competitive semantic performance and high consistency. The complete pipeline demonstrates a proof-of-concept methodology using Hindi for developing specialized AI systems for any languages with WordNet resources. This work addresses the critical gap in AI accessibility for low-resource languages, offering a practical alternative to corpus-intensive approaches and potentially enabling specialized AI development for the hundreds of languages with existing WordNet resources.
 
\end{abstract}

\section{Introduction}

The democratization of artificial intelligence increasingly depends on developing specialized systems that effectively serve diverse linguistic communities. While recent advances in large language models have shown strong capabilities \citep{10638808}, these systems predominantly excel in high-resource languages with abundant digital content, leaving billions of speakers of low-resource languages underserved \citep{zhong2024opportunitieschallengeslargelanguage, Hasan2024}. This gap is particularly pronounced in specialized domains such as education, where culturally and linguistically aligned AI systems are critical for effective learning outcomes \citep{Zhang2024}.

A key bottleneck in developing conversational AI for such settings is the reliance on massive training corpora, which are unavailable for most of the world’s 2,500+ languages \citep{UNESCO2010, EndangeredLanguagesProject}. Existing fine-tuning paradigms assume large-scale textual data \citep{cryst2025mind}, as seen in datasets used for GPT-3 \citep{NEURIPS2020_1457c0d6} and Common Crawl \citep{Baack2024CommonCrawl}. However, the challenge extends beyond data quantity to the lack of high-quality, domain-specific instructional data. Even languages with substantial web presence, such as Hindi, remain “resource-poor” in terms of pedagogically structured examples, limiting their applicability in specialized AI systems.

To address this limitation, we leverage structured linguistic resources that encode expert-curated knowledge. WordNets which are hierarchical lexical databases capturing semantic relationships, exist for over 200 languages \citep{GlobalWordNetAssociation}, while resources like BabelNet integrate multilingual WordNets at scale \citep{Navigli2010}. Despite their richness, such resources are typically used as static references rather than as foundations for training conversational systems, representing a missed opportunity for advancing AI accessibility in multilingual contexts.

We propose a systematic methodology for transforming structured linguistic resources into specialized conversational AI systems, providing a practical alternative to corpus-intensive approaches. Using Hindi WordNet \citep{bhattacharyya-2010-indowordnet, HindiWordNetLDC2008} as a case study, we construct a language-learning system and evaluate it across semantic accuracy and pedagogical effectiveness.

Beyond this specific implementation, our approach suggests a potentially scalable pathway for enabling AI development across languages with existing structured resources. Hindi WordNet, comprising 105,460 words and 40,466 synsets \citep{HindiWordNet2025Website}, also serves as a foundation for other Indian language WordNets. This makes the methodology particularly relevant for educational applications in resource-constrained settings, where both computational limitations and lack of appropriate content hinder adoption \citep{Redkar2018}.

Our key contributions are: (1) a systematic method for converting structured lexical databases into diverse conversational training data while preserving semantic relationships; (2) demonstration off a resource-efficient fine-tuning framework enabling deployment in typical educational environments; and (3) empirical evidence demonstrating improved domain-specific performance over general-purpose models. Together, these results establish a reproducible pipeline for building specialized AI systems in low-resource languages.

\section{Related Work}

\subsection{The Evolving Role of AI in Education}
Large-scale surveys consistently report positive learning gains from AI interventions while warning that impact is often measured on single dimensions rather than intertwined pedagogical, technical, and human factors. A comprehensive review covering 2010-2020 recommends "a multidimensional evaluation model" combining technical metrics with pedagogical design, domain alignment, and learner affect \citep{Zhai2021}. A conceptual synthesis categorizes AI's functions into three roles: new subject, direct mediator, and supplementary assistant-showing how each reshapes classroom dynamics \citep{Xu2022}. When AI takes the "new subject" role (e.g., tutoring agent), it can personalize instruction but must address social presence and reflection to avoid merely automating drill-and-practice \citep{Xu2022}. These insights frame our approach as maintaining learner connections to structured knowledge rather than replacing expert \looseness=-1 guidance.

\subsection{Chatbots for Language Learning}
Systematic evidence confirms three recurring affordances of language-learning chatbots: timeliness, ease of access, and personalization, with pedagogical uses including simulation, helpline, and recommendation \citep{Huang2021}. Social-presence analyses show bot self-disclosure encourages longer learner utterances and reduces practice anxiety \citep{Huang2021}. CLIL field studies demonstrate high engagement (91\% content mastery agreement, 93\% finding dialogue engaging) but only 48\% felt language skill improvement, highlighting content-language objective tensions \citep{Mageira2022}. These findings motivate our level-adaptive output balancing vocabulary complexity with curricular content, and post-response augmentation sustaining engagement beyond novelty effects.

\subsection{Conversational AI in Low-Resource Languages}
Low-resource contexts add data scarcity, cultural nuance, and deployment constraints to AI development challenges. Vision papers argue techniques like Direct Preference Optimization can lower supervision requirements for culturally sensitive AI companions \citep{Ding2024}. Empirical work explores lightweight architectures: a Bangla customer service bot achieves >90\% accuracy using n-gram stemming and CNN classifiers without deep linguistic resources, but lacks structured knowledge integration and level adaptation \citep{Paul2019}. Knowledge-enriched FAQ chatbots improve intent classification through transfer learning but rely on retrieval rather than generation, limiting conversational depth \citep{Perdana2021}. More recent approaches attempt to bypass data scarcity by leveraging the cross-lingual transfer capabilities of large foundation models. \citep{nguyen-etal-2024-democratizing} introduced Linguistically-Diverse Prompting (LDP), a technique that elicits generative capabilities in low-resource languages by using synthetic exemplars from high-resource "sibling" languages or English pivots. While this method demonstrates that English-dominant models can be coaxed into performing translation and summarization tasks for under-represented languages without supervised data, it fundamentally relies on the model's latent, pre-trained knowledge. Consequently, such prompting strategies remain susceptible to the hallucination and factual inconsistency inherent in the base model, particularly in specialized domains where the model's internal representation of the low-resource language is sparse or fragmented. This limitation underscores the necessity for methods that can ground generation in explicit, structured expert knowledge rather than relying solely on cross-lingual transfer. These studies demonstrate feasibility while underscoring gaps: (1) automatic diverse instruction-response generation; (2) resource-efficient fine-tuning; and (3) structured lexical resource coupling \looseness=-1 \citep{Oyewole2023}.

Hindi WordNet has been adapted into Hindi Shabdamitra, a five-level digital aid exposing gloss simplification and progressively richer semantic relations to K-12 learners. Classroom pilots show improved concept retention when learners explore associative networks rather than flat dictionary entries \citep{Redkar2018}. WordNet's cognitive basis, which represents meaning as concept networks, aligns with semantic network vocabulary acquisition theories. Despite this potential, existing conversational systems rarely exploit such structure beyond initial training. Our approach bridges this gap by converting synsets into training examples, maintaining knowledge connections through post-generation augmentation, and enabling conversation-to-structure pivoting.

\subsection{Research Gaps and Opportunities}
Critical gaps remain for low-resource language applications: (1) Structured knowledge continuity - chatbots rarely maintain learner connections to training resources \citep{Huang2021, Oyewole2023}; (2) Level-adaptive generation - few systems systematically vary vocabulary, syntax, and explanation depth across proficiency levels \citep{Paul2019}; (3) Resource-efficient deployment - approaches often assume cloud-scale hardware \citep{Ding2024}; and (4) Integrated scaffolding - studies report novelty effects and limited long-term gains, indicating the need for dynamic learning supports \citep{Mageira2022, Huang2021}. Our methodology addresses each gap by coupling structured linguistic resources with parameter-efficient fine-tuning and real-time knowledge augmentation.

\section{Methodology: Structured-Data-to-AI Pipeline}

Our systematic methodology transforms structured linguistic databases into specialized conversational AI systems through four integrated stages: systematic dataset generation, resource-efficient model fine-tuning, domain-adaptive response generation, and intelligent knowledge integration. This pipeline suggests that expert-curated lexical resources can serve as useful foundations for specialized AI development, offering a practical alternative to corpus-intensive approaches for low-resource languages.

\subsection{Dataset Creation Pipeline}
\subsubsection{Structured Knowledge Processing}
We systematically convert Hindi WordNet's structured semantic data into diverse conversational training examples. The resource contains 56,928 words with rich semantic relationships including hypernymy, hyponymy, meronymy, antonymy, and ontological hierarchies. Our automated pipeline generates four complementary types of instruction-response pairs designed to preserve the structured knowledge while creating natural conversational interactions (see Appendix \ref{sec:appendix_examples} for detailed examples):

\begin{itemize}[leftmargin=*, noitemsep, topsep=2pt]
    \item \textbf{Basic Instructional Pairs} establish fundamental question-answer patterns for core linguistic concepts.
    \item \textbf{Complex Multi-Aspect Pairs} integrate multiple semantic relationships within single responses, teaching comprehensive word understanding including definitions, synonyms, examples, and grammatical categories within complete linguistic contexts. 
    \item \textbf{Ontological Hierarchy Pairs} leverage WordNet's taxonomic structure to teach categorical relationships.
    \item \textbf{Disambiguation Pairs} address polysemy by explicitly teaching multiple word meanings with contextual differentiation, crucial for morphologically rich languages like Hindi.
\end{itemize}

\subsubsection{Relational Coverage and Quality Assurance}
Hindi WordNet's 23 semantic relationship types \citep{bhattacharyya-2010-indowordnet} are mapped to educational terminology through expert linguistic consultation. To prevent information overload while ensuring complete coverage, we implemented a dynamic chunking algorithm for dense semantic relationships. For entries with extensive relational lists (exceeding 10 related terms), our algorithm splits the data into digestible segments of up to 10 items. Crucially, we enforce a 33\% overlap (approx. 3 words) between consecutive chunks. This overlapping sliding-window strategy ensures the model learns the continuity of semantic categories across training examples rather than treating them as disjoint sets.

Furthermore, to foster multi-hop semantic reasoning, the pipeline dynamically combines distinct relationship types into unified complex queries. For example, a single instruction might require the model to identify both the \textit{Hypernym} and \textit{Antonym} of a target word. We implemented a coverage check to ensure that every unique word in a relation list appears in at least one training example, whether in a single or combined query. Finally, the pipeline implements hash based deduplication using an instruction-response hash comparison, removing 847,000 duplicate examples from an initial 2.1 million generated pairs. The final dataset consists of 1,253,847 unique pairs with balanced representation across all relationship types.

\subsection{Resource-Efficient Model Specialization}
\subsubsection{Base Model Selection and Optimization}
We select Gemma-3-12B-IT as our foundation model for its demonstrated multilingual capabilities and instruction-following performance \citep{GemmaTeam2025}. To enable deployment in resource-constrained environments, we implemented 4-bit quantization using NF4 (Normalized Float 4) with double quantization \citep{Dettmers2023}, reducing memory requirements from 48GB to approximately 12GB while preserving model performance - a critical consideration for low-resource language applications where computational resources are limited.

\subsubsection{Parameter-Efficient Fine-Tuning Configuration}

We specialize the model using Low-Rank Adaptation (LoRA) \citep{Hu2022} with optimized hyperparameters balancing adaptation capability with efficiency: \textbf{Rank (r)}: 32, providing sufficient expressiveness for domain specialization; \textbf{Alpha}: 64, ensuring appropriate scaling for knowledge adaptation; \textbf{Target modules}: all attention projections and MLP components for comprehensive adaptation; and \textbf{Dropout}: 0.05 for regularization without overfitting. This configuration fine-tunes only 0.2\% of the total parameters (67M out of 12B), enabling rapid specialization while preserving pre-trained multilingual knowledge, essential for maintaining general linguistic competence during domain adaptation.

\subsubsection{Training Configuration and Efficiency}
Our training employs distributed setup with gradient accumulation achieving effective batch sizes of 8 across available hardware. Key parameters include 2e-5 learning rate with cosine scheduling \citep{Loshchilov2017}, 15\% warmup steps, gradient clipping at 0.5 for stability \citep{Pascanu2013}, and 3 training epochs with early stopping. The complete training process requires approximately 40 hours on 2×NVIDIA A100 80G GPUs, demonstrating practical feasibility for educational institutions and research organizations in developing regions.

\subsection{Domain-Adaptive Response Generation}

We implement systematic level adaptation aligned with educational curricula through structured prompt templates defining five proficiency levels - from \texthindi{प्राथमिक} (Beginner, 2--3 simple sentences with everyday vocabulary and concrete examples), \texthindi{माध्यमिक} (Intermediate, 4--5 sentences with practical examples), \texthindi{कुशल} (Proficient, 6--8 sentences with increased grammatical nuance and varied examples), \texthindi{उन्नत} (Advanced, 8--10 sentences incorporating abstract and cultural context), to \texthindi{विशेषज्ञ} (Expert, 10+ sentences with technical linguistic terminology and interdisciplinary analysis). Each template explicitly controls vocabulary complexity, sentence length, explanation depth, and example types, enabling consistent and reproducible response adaptation across proficiency levels.

Given the educational deployment context targeting children, safety constraints are embedded directly into the base system prompt, restricting responses to family, school, and nature domains while blocking inappropriate, violent, or adult content, with fallback mechanisms for unsafe queries \citep{Gehman2020}. Full prompt templates for all five levels are provided in Appendix \ref{sec:appendix_prompts}.

We refer to this fine-tuned Gemma-3-12B-IT model as \textit{Shabdabot} throughout the remainder of the paper.

\section{Results}

\subsection{Evaluation Setup and Metrics}
We conducted a rigorous comparative evaluation using 40 carefully designed Hindi language questions spanning five proficiency levels (\texthindi{प्राथमिक} (Beginner) to \texthindi{विशेषज्ञ} (Expert)). Expert linguists created golden reference answers for each question-level combination, resulting in 200 reference responses. We obtained responses from five models- Shabdabot, GPT-4.1 \citep{OpenAI2025}, Claude-Sonnet-4 \citep{Anthropic2025}, Gemini-2.5Pro \citep{Google2025}, and Gemma-3-12B-IT \citep{GemmaTeam2025}-using identical prompts and system settings to ensure fair comparison.

To eliminate evaluation bias, all responses were anonymized during metric calculation. We employed two complementary evaluation metrics designed to assess both semantic accuracy and pedagogical effectiveness:

\textbf{Semantic Answer Similarity (SAS)} measures the semantic fidelity between model responses and expert-created golden answers. This metric employs the multilingual sentence transformer \textit{paraphrase-multilingual-MiniLM-L12-v2} \citep{Reimers2020} to generate vector embeddings for both model responses and reference answers. Semantic similarity is calculated using cosine similarity between these embeddings, producing scores ranging from 0 to 1, where higher values indicate greater semantic alignment with expert-authored content. The multilingual model was specifically chosen for its demonstrated effectiveness in cross-lingual semantic similarity tasks and strong performance on Hindi text. This metric captures how well models preserve the core meaning and factual content of expert responses, independent of stylistic or pedagogical considerations.

\textbf{Level Adaptation Quality (LAQ)} assesses pedagogical effectiveness and appropriateness for educational contexts through expert evaluation. We employed Claude-Sonnet-4 as an automated expert judge, chosen for its demonstrated reliability in educational content evaluation and ability to process Hindi text with cultural and linguistic nuance. The LAQ evaluation employs a comprehensive rubric that evaluates five pedagogical criteria: (1) Pedagogical Clarity - how easily the target learner can understand the explanation; (2) Factual accuracy - correctness and precision of provided information; (3) Relevance \& Examples-appropriateness and quality of examples for the proficiency level; (4) Language Appropriateness-suitability of vocabulary, syntax, and tone for the intended learner; and (5) Educational Value - general utility as a teaching tool for the specific proficiency level. Each criterion receives a score from 0-20 points, yielding total scores from 0-100, with higher scores indicating superior educational effectiveness. To ensure evaluation reliability, we provided detailed scoring rubrics with level-specific criteria and conducted consistency validation across multiple evaluation runs.

These complementary metrics enable comprehensive assessment of both semantic competence and domain-specific effectiveness, addressing the critical question of whether specialized systems can maintain linguistic accuracy while achieving superior pedagogical outcomes compared to general-purpose models.

To validate the reliability of this automated judging, we conducted a human-in-the-loop verification process. A team of expert Hindi linguists reviewed a randomized subset of the model responses and the corresponding scores assigned by Claude-Sonnet-4. The experts verified that the AI judge consistently applied the rubric criteria regarding all five pedagogical criteria. This expert cross-verification confirmed that the automated scores were not random and aligned closely with human pedagogical assessment standards, validating the use of the LLM as a scalable proxy for expert evaluation.

\begin{table*}[t]
\centering
\begin{tabular}{lccccc}
\toprule
\textbf{Model} & \textbf{LAQ Score} & \textbf{LAQ Rank} & \textbf{Consistency ($\sigma$)} & \textbf{SAS Score} & \textbf{SAS Rank} \\
\midrule
Shabdabot        & \textbf{91.0} & \textbf{1st} & \textbf{1.0} & 0.731 & 2nd \\
GPT-4.1          & 79.4 & 5th & 7.4 & \textbf{0.762} & \textbf{1st} \\
Gemma-3-12B-IT   & 80.8 & 4th & 2.4 & 0.728 & 3rd \\
Claude-Sonnet-4  & 81.9 & 3rd & 6.4 & 0.712 & 4th \\
Gemini-2.5Pro    & 83.6 & 2nd & 5.7 & 0.705 & 5th \\
\bottomrule
\end{tabular}
\caption{Overall Model Performance}
\label{tab:overall_performance}
\end{table*}

% \begin{table*}[t]
% \centering
% \label{tab:sas_by_level}
% \begin{tabular}{lccccc}
% \toprule
% \textbf{Level} & \textbf{Shabdabot} & \textbf{GPT-4.1} & \textbf{Gemma-3-12B-IT} & \textbf{Claude-Sonnet-4} & \textbf{Gemini-2.5Pro} \\
% \midrule
% \texthindi{प्राथमिक} & 0.714 & \textbf{0.736} & 0.721 & 0.683 & 0.728 \\
% \texthindi{माध्यमिक} & 0.753 & \textbf{0.793} & 0.747 & 0.748 & 0.781 \\
% \texthindi{कुशल}     & 0.741 & \textbf{0.791} & 0.748 & 0.761 & 0.763 \\
% \texthindi{उन्नत}       & \textbf{0.759} & 0.757 & 0.737 & 0.703 & 0.691 \\
% \texthindi{विशेषज्ञ}  & 0.688 & \textbf{0.735} & 0.686 & 0.663 & 0.563 \\
% \bottomrule
% \end{tabular}
% \caption{SAS Performance by Proficiency Level}
% \end{table*}

\subsection{Overall Performance Analysis}
The results reveal a critical insight for specialized AI development: while GPT-4.1 achieved highest semantic similarity to expert-created answers, our structured-knowledge-based system dramatically outperformed all models in domain-specific effectiveness with a 91.0 LAQ score-an 11.6-point advantage over the second-best model and a remarkable 12.6\% improvement over its base model (see Table \ref{tab:overall_performance}).

\subsection{Proficiency Level Performance Patterns}
\begin{figure}[!t]
\centering
\includegraphics[width=\columnwidth]{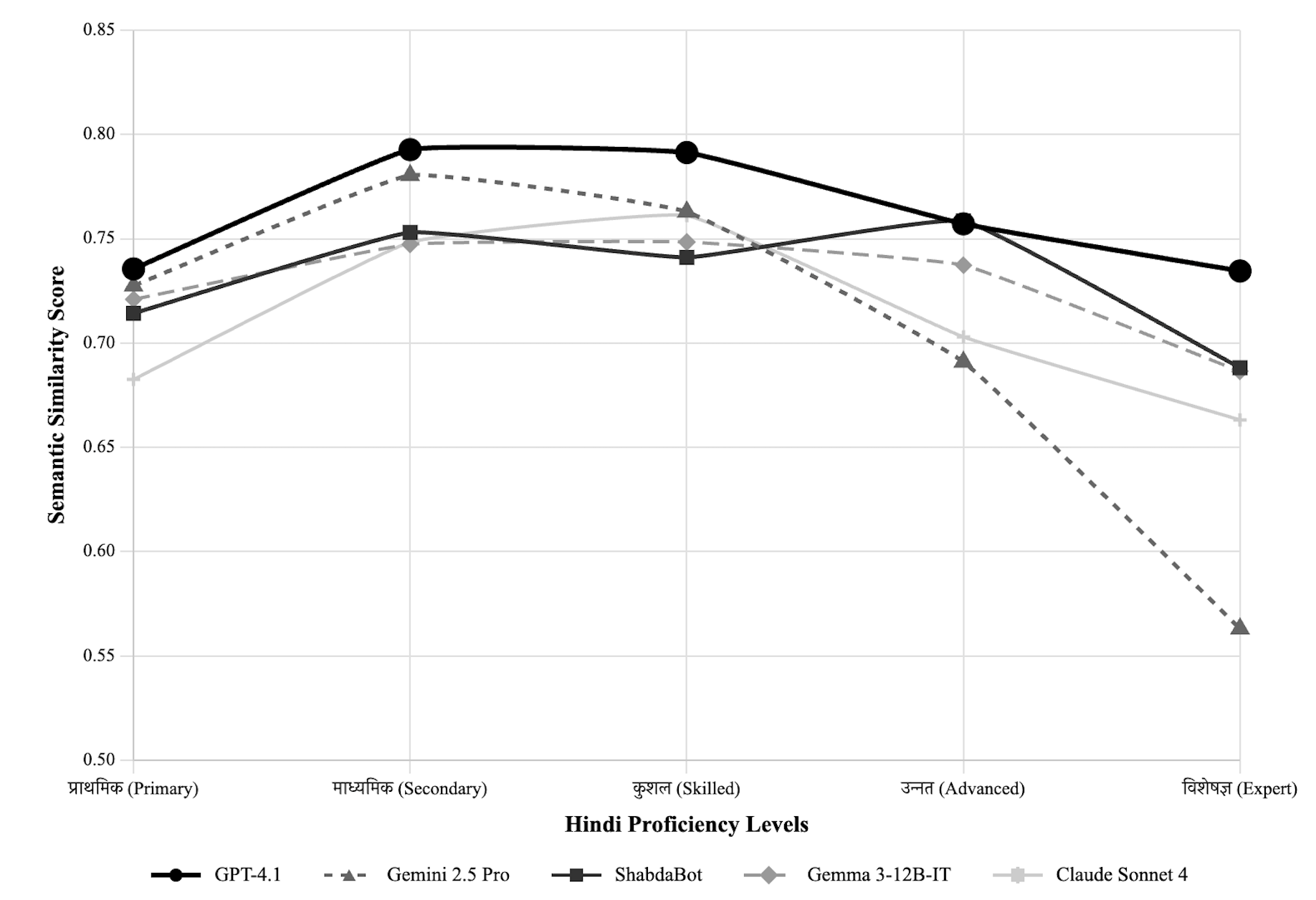}
\caption{SAS Analysis visualization across proficiency levels.}
\label{fig:sas_analysis}
\end{figure}

Figure \ref{fig:sas_analysis} illustrates semantic performance across proficiency levels, revealing distinct patterns supporting our structured-knowledge approach. %Table \ref{tab:sas_by_level} provides the detailed scores.

\textbf{Critical Finding:} Shabdabot uniquely peaks at the \texthindi{उन्नत} (Advanced) level, achieving highest performance among all models at this level. The performance decline at the \texthindi{विशेषज्ञ} (Expert) level reflects training data characteristics-our structured-knowledge conversion emphasized educational clarity over lengthy academic discourse typical of expert-level responses.

The LAQ evaluation shows Shabdabot's consistent performance across all proficiency levels:
\begin{itemize}[noitemsep, topsep=0pt]
    \item \textbf{Primary to Expert levels:} 83.0-83.8 (standard deviation: 0.37)
    \item \textbf{Best performer:} All five proficiency levels
    \item \textbf{Stability:} Unlike general-purpose models showing significant performance degradation with difficulty increases
\end{itemize}

\subsection{Statistical Significance and Reliability Analysis}
\begin{figure}[!t]
\centering
\includegraphics[width=\columnwidth]{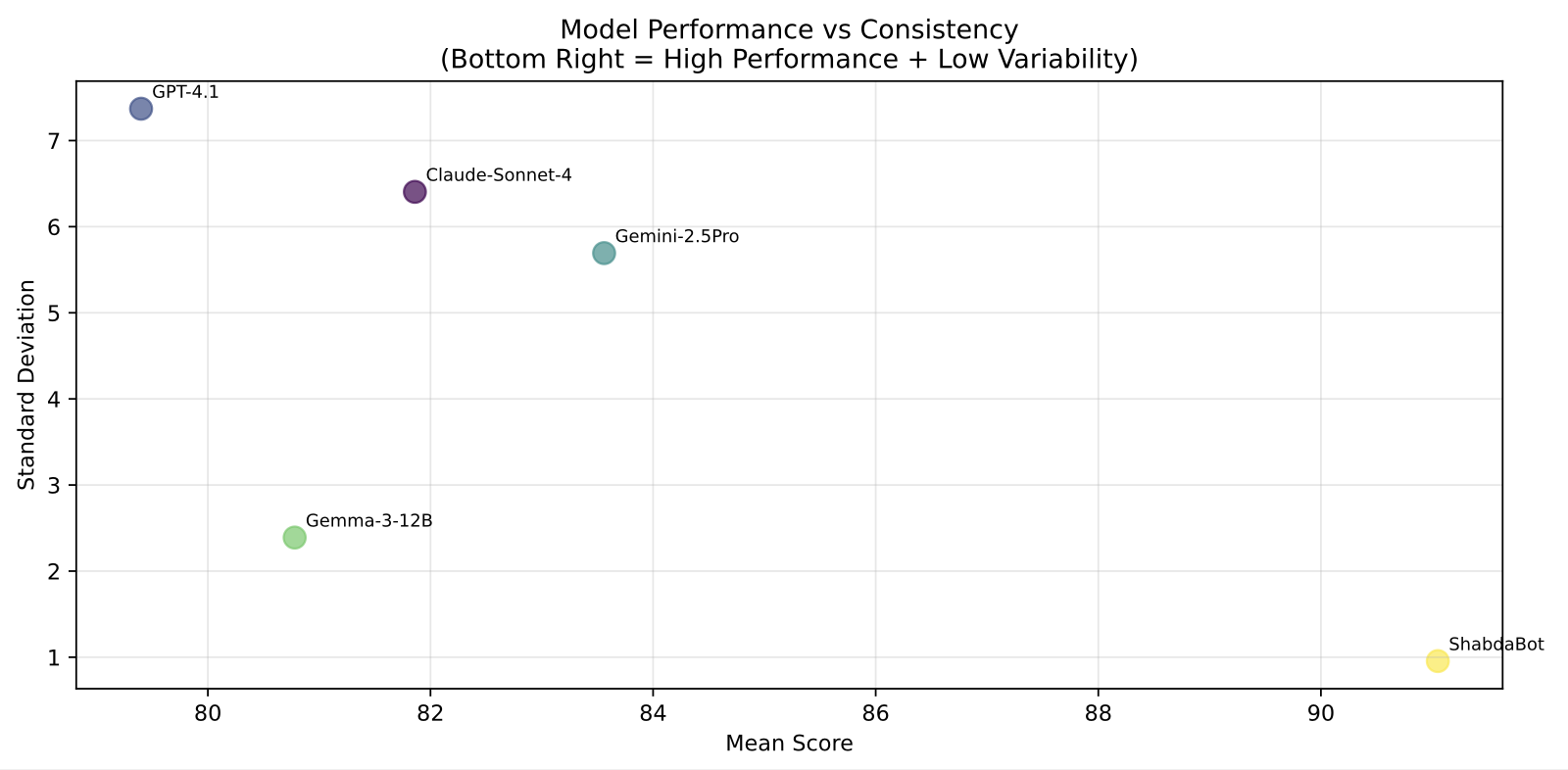}
\caption{Performance vs. Consistency scatter plot.}
\label{fig:consistency_plot}
\end{figure}

One-way ANOVA confirmed significant differences between models ($F(4,995) = 5.491, p < 0.001$). Key findings include:
\begin{itemize}[noitemsep, topsep=0pt]
    \item \textbf{Semantic Performance:} GPT-4.1 vs. Shabdabot significant difference ($p = 0.019$, Cohen's d = 0.236) \citep{Diener2010}, while Shabdabot vs. Gemma-3-12B-IT showed non-significant difference ($p = 0.819$), indicating preserved semantic competence during specialization.
    \item \textbf{Reliability Advantage:} Figure \ref{fig:consistency_plot} highlights a key advantage of our approach-Shabdabot achieved high consistency with $\sigma = 1.0$ compared to 7.4 for GPT-4.1, representing an 86\% improvement in predictability. Reliability metrics are detailed in Table \ref{tab:reliability_metrics}.
\end{itemize}

\begin{table}[!htbp]
\centering
\begin{tabular}{lccc}
\toprule
\textbf{Model} & \shortstack{\textbf{LAQ}\\\textbf{Std Dev}} &  \shortstack{\textbf{High Perf.}\\\textbf{(>90\%)}} \\
\midrule
Shabdabot        & \textbf{1.0} & \textbf{93\%} \\
Gemma-3-12B-IT   & 2.4  & 0\% \\
Gemini-2.5Pro    & 5.7  & 6\% \\
Claude-Sonnet-4  & 6.4  & 5.5\% \\
GPT-4.1          & 7.4 & 0\% \\
\bottomrule
\end{tabular}
\caption{Reliability Metrics Comparison}
\label{tab:reliability_metrics}
\end{table}

\subsection{Domain Specialization Effectiveness}

\begin{figure}[!t]
\centering
\includegraphics[width=\columnwidth]{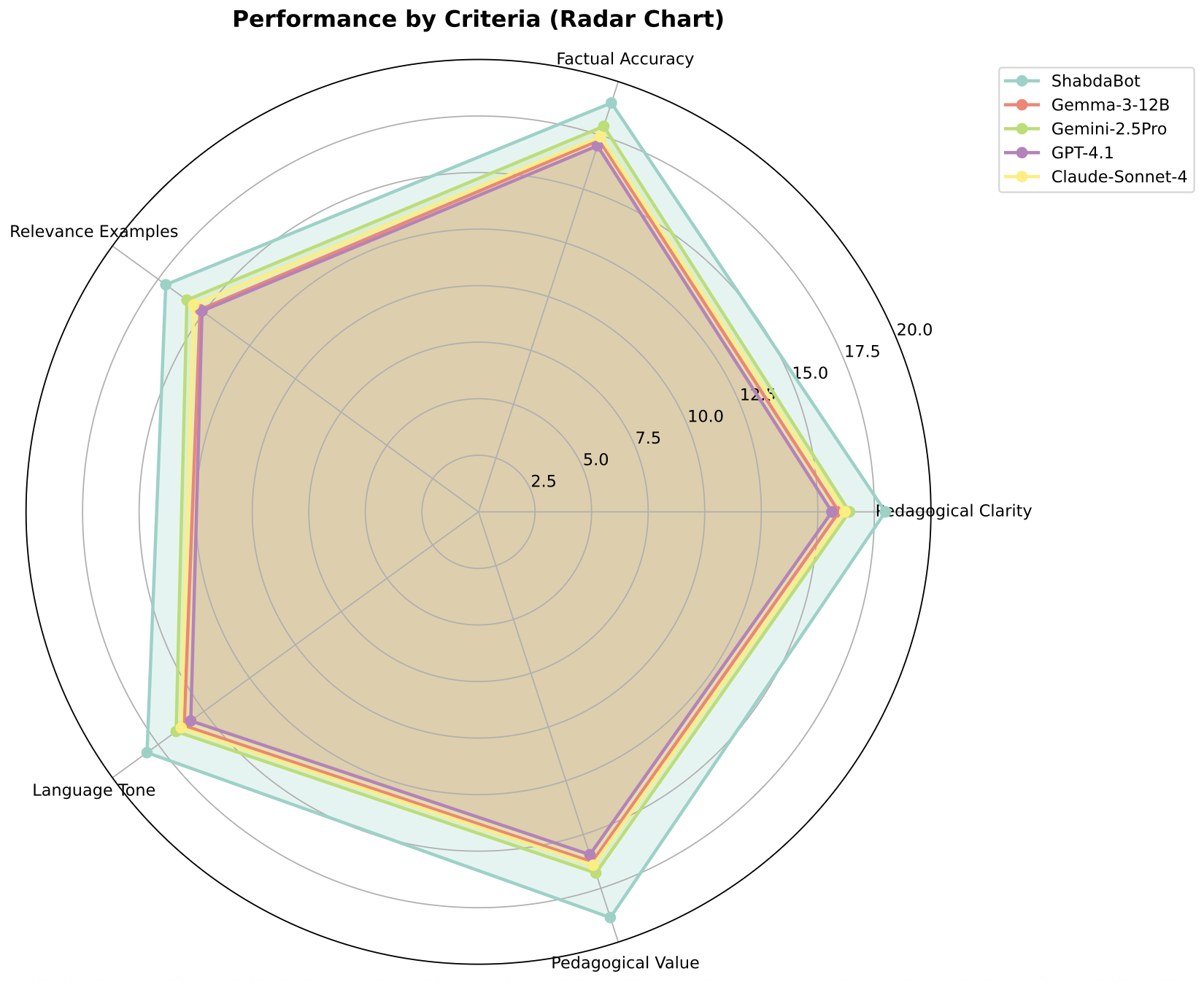}
\caption{Radar Chart of Performance Across Pedagogical Criteria.}
\label{fig:radar_chart}
\end{figure}

Figure \ref{fig:radar_chart} shows that Shabdabot received higher pedagogical scores across all pedagogical criteria. The model achieved high scores in Pedagogical Clarity (18.2/20), Language Appropriateness (18.4/20), Relevance \& Examples (18.1/20), and Educational Value (17.8/20), while maintaining competitive Factual Accuracy (18.5/20).

\subsection{Structured-Knowledge Impact Analysis}
Here, \textit{domain effectiveness} refers to the Level Adaptation Quality (LAQ) score (0--100), our primary metric for pedagogical quality in the target educational domain. Direct comparison between our system and its base model reveals the effectiveness of structured-knowledge specialization, as shown in Table \ref{tab:specialization_impact}.

\begin{table}[!htbp]
\centering
\begin{tabular}{l@{\hspace{4pt}}c@{\hspace{4pt}}c@{\hspace{4pt}}c}
\toprule
\textbf{Metric} & \shortstack{\textbf{Gemma-3}\\[0pt]\textbf{12B-IT}} & \textbf{Shabdabot} & \shortstack{\textbf{Improve-}\\[0pt]\textbf{ment}} \\
\midrule
\shortstack{Semantic\\[0pt]Comp.} & 0.728 & 0.731 & \bf{+0.4\%} \\
\shortstack{Domain\\[0pt]Eff.(LAQ)} & 80.8 & 91.0 & \bf{+12.6\%} \\
Consistency & 2.4$\sigma$ & 1.0$\sigma$ & \bf{+58\%} \\
\shortstack{Advanced\\[0pt]SAS} & 0.737 & 0.759 & \bf{+2.9\%} \\
\shortstack{Educ.\\[0pt]Failures} & 0 & 0 & Maintained \\
\bottomrule
\end{tabular}
\caption{Specialization Impact}
\label{tab:specialization_impact}
\end{table}

Direct comparison between our system and its base model reveals the effectiveness of structured-knowledge specialization, as shown in Table \ref{tab:specialization_impact}. Domain effectiveness improves by 12.6\% while semantic competence is preserved, and response consistency improves by 58\% ($\sigma: 2.4 \to  1.0$) - suggesting that structured knowledge integration enhances rather than constrains general linguistic capability, while producing a more predictable and reliable system.

Together, these results confirm that domain specialization and practical deployability are not in tension - structured knowledge integration achieves both simultaneously.

\section{Discussions}
The performance gap between Shabdabot and general-purpose models reveals a counterintuitive insight: semantic similarity to expert answers does not predict pedagogical effectiveness in specialized domains. The results show important patterns: while general-purpose models like GPT-4.1 achieve higher semantic similarity to expert-created responses, our structured-knowledge-based system shows higher domain-specific effectiveness. This challenges the prevailing assumption that general-purpose foundation models are inherently optimal for specialized, pedagogical applications.

The high consistency achieved by our approach-an 86\% improvement in reliability compared to leading general-purpose models-addresses practical concerns for AI deployment, particularly in educational contexts where unpredictable responses can confuse learners. We hypothesize that this reliability stems from systematic structured knowledge integration, where generation is grounded in expert-curated linguistic relationships rather than statistical patterns derived from unstructured web text. By demonstrating that 1.25 million structured examples can create specialized systems that can be competitive with models trained on billions of general examples, we establish that expert-curated knowledge can effectively complement or substitute for corpus-intensive approaches.

Several limitations suggest important considerations for future applications. Expert-level performance decline points to a training data gap we detail in Section 6. Evaluation of long-term learning outcomes in authentic classrooms remains an important open question. Cross-linguistic validation beyond Hindi remains necessary to establish generalizability.

\section{Conclusion}
This work reframes how we think about data for low-resource AI development. The prevailing assumption is that more general-purpose data produces better systems - our results directly challenge this, showing that 1.25 million wordnet generated structured examples outperform models trained on orders of magnitude more unstructured text, specifically in the domain they are designed for. The consistency gap is particularly telling: Shabdabot's σ=1.0 against GPT-4.1's σ=7.4 suggests that grounding generation in structured knowledge doesn't just improve accuracy - it fundamentally changes the reliability profile of the system.

Crucially, the methodology's resource efficiency-achieving superior domain performance while requiring only 12GB RAM-directly addresses the computational barriers that prevent communities from accessing sophisticated AI technologies. With WordNets and similar structured resources available for over 200 languages, this pipeline offers a highly reproducible development pathway. Ultimately, this work provides a practical framework for democratizing AI, leveraging decades of linguistic scholarship to deliver equitable, domain-specific systems for billions of underserved language speakers worldwide.

\section{Limitations}

While our results demonstrate the effectiveness of structured-knowledge approaches for specialized AI development, several limitations warrant consideration:

\textbf{Training Data Coverage}: Our automated pipeline emphasized educational clarity and conciseness, potentially underrepresenting the verbose, technically dense responses characteristic of expert-level academic discourse. This is evident in the Expert-level SAS score drop to 0.688 - the lowest across all levels - compared to 0.759 at Advanced. Future pipelines could address this by supplementing WordNet with academic corpora or long-form Wikipedia text specifically for expert-level examples.

\textbf{Dependence on Structured Resource Availability.}
A limitation of our approach is its dependence on the availability and quality of structured linguistic resources. While WordNets and similar databases exist for many languages, their coverage, granularity, and domain completeness vary significantly, and may be limited or absent for truly low-resource languages. In such cases, additional effort is required to construct or augment these resources, which introduces overhead in adopting the methodology. However, compared to corpus-intensive approaches, structured resources are significantly more compact and reusable, making them a promising foundation for scalable multilingual AI development once such resources are available.

\textbf{Domain and Language Scope}: Our evaluation focuses exclusively on Hindi language education. Specifically, agglutinative languages like Tamil or Turkish, and isolating languages like Mandarin, present structurally different WordNet organizations that may require pipeline modifications - this remains untested.

\textbf{Long-term Impact Assessment}: Our evaluation measures immediate response quality and pedagogical appropriateness rather than actual learning outcomes. Longitudinal studies tracking learner vocabulary retention over 4–8 weeks, using pre/post assessments, would provide the most meaningful validation of actual educational impact beyond our current response-quality metrics.

These limitations suggest important directions for future work while not diminishing the core contribution of demonstrating that structured linguistic resources can effectively serve as foundations for specialized AI development in low-resource language contexts.

\bibliography{custom}
\bibliographystyle{acl_natbib}

\appendix

\section{Prompt Engineering Templates}
\label{sec:appendix_prompts}

This appendix outlines the specific system instructions and level-adaptive prompts used in the generation pipeline. All prompts were administered in Hindi to ensure linguistic consistency.

\subsection{Base System Instruction}
All proficiency levels shared the following core system prompt, which establishes the persona, scope, and safety constraints:

\begin{promptbox}[Base System Prompt]
\texthindi{आप एक हिंदी भाषा शिक्षण चैटबॉट हैं जिसका एकमात्र उद्देश्य बच्चों को हिंदी भाषा संबंधी शैक्षिक जानकारी प्रदान करना है। कृपया नीचे दिए गए नियमों का कड़ाई से पालन करें:}

\begin{enumerate}[leftmargin=*]
    \item \texthindi{केवल हिंदी भाषा के अर्थ, व्याकरण, समानार्थी, विलोम, पराजाति-अपराजाति, अवयव-अवयवी संबंध, वाक्य में शब्दों का प्रयोग, और पदानुक्रम} (ontology) \texthindi{संबंधी प्रश्नों का उत्तर दें।}
    \item \texthindi{आपकी जानकारी सरल, स्पष्ट, सकारात्मक, और बच्चों के लिए पूरी तरह सुरक्षित होनी चाहिए।}
    \item \texthindi{उदाहरण केवल परिवार, स्कूल, घर, प्रकृति, पशु-पक्षी, मित्रता और सकारात्मक गतिविधियों तक सीमित रखें।}
    \item \texthindi{मित्रता या परिवार से बाहर के संबंधों के संदर्भ को गलत समझे जाने वाले उदाहरण न दें।}
    \item \texthindi{अनुचित, हिंसक, डरावनी, वयस्क, या अश्लील सामग्री का उल्लेख किसी भी परिस्थिति में न करें।}
\end{enumerate}
\end{promptbox}

\begin{promptbox}[Base System Prompt (English Translation)]
You are a Hindi language teaching chatbot whose sole purpose is to provide children with educational information related to the Hindi language. Please strictly follow the rules given below:

\begin{enumerate}[leftmargin=*]
    \item Answer only questions related to Hindi language meaning, grammar, synonyms, antonyms, hypernym-hyponym relations, meronym-holonym relations, word usage in sentences, and ontology hierarchy.
    \item Your information must be simple, clear, positive, and completely safe for children.
    \item Keep examples limited only to family, school, home, nature, animals and birds, friendship, and positive activities.
    \item Do not provide examples that could be misinterpreted in the context of relationships outside friendship or family.
    \item Do not mention inappropriate, violent, frightening, adult, or obscene content under any circumstances.
\end{enumerate}
\end{promptbox}

\subsection{Proficiency Level Instructions}
Specific instructions were injected dynamically based on the target proficiency level.

\paragraph{Level 1: \texthindi{प्राथमिक} (Beginner)}
\begin{promptbox}
\texthindi{आप एक सहायक हिंदी शिक्षक हैं। उत्तर बिल्कुल सरल भाषा में दें:}
\begin{itemize}[leftmargin=*, noitemsep]
    \item \texthindi{बहुत आसान शब्दों का प्रयोग करें}
    \item \texthindi{छोटे वाक्य बनाएं}
    \item \texthindi{यदि जरूरी हो तो अंग्रेजी शब्द का भी प्रयोग कर सकते हैं}
    \item \texthindi{उदाहरण रोजमर्रा की जिंदगी से दें}
    \item \texthindi{उत्तर संक्षिप्त रखें (2-3 वाक्य)}
    \item \texthindi{बच्चों को समझ आने वाली भाषा का प्रयोग करें}
\end{itemize}
\end{promptbox}

\paragraph{Level 1: Beginner}
\begin{promptbox}
You are a helpful Hindi teacher. Please answer in very simple language:
\begin{itemize}[leftmargin=*, noitemsep]
    \item Use very easy words.
    \item Make short sentences.
    \item You can also use English words if necessary.
    \item Give examples from everyday life.
    \item Keep the answer brief (2-3 sentences).
    \item Use language that children can understand.
\end{itemize}
\end{promptbox}

\paragraph{Level 2: \texthindi{माध्यमिक} (Intermediate)}
\begin{promptbox}
\texthindi{आप एक अनुभवी हिंदी शिक्षक हैं। उत्तर मध्यम स्तर की भाषा में दें:}
\begin{itemize}[leftmargin=*, noitemsep]
    \item \texthindi{सामान्य हिंदी शब्दावली का प्रयोग करें}
    \item \texthindi{मध्यम लंबाई के वाक्य बनाएं}
    \item \texthindi{व्यावहारिक उदाहरण दें}
    \item \texthindi{मुख्य बिंदुओं को स्पष्ट रूप से समझाएं}
    \item \texthindi{उत्तर 4-5 वाक्यों में दें}
    \item \texthindi{विषय की बुनियादी जानकारी प्रदान करें}
\end{itemize}
\end{promptbox}

\paragraph{Level 2: Intermediate}
\begin{promptbox}
You are an experienced Hindi teacher. Please answer in intermediate-level language:
\begin{itemize}[leftmargin=*, noitemsep]
    \item Use common Hindi vocabulary.
    \item Make medium-length sentences.
    \item Give practical examples.
    \item Explain the main points clearly.
    \item Provide the answer in 4-5 sentences.
    \item Provide basic information about the topic.
\end{itemize}
\end{promptbox}

\paragraph{Level 3: \texthindi{कुशल} (Proficient)}
\begin{promptbox}
\texthindi{आप एक कुशल हिंदी भाषा विशेषज्ञ हैं। उत्तर अच्छी गुणवत्ता की भाषा में दें:}
\begin{itemize}[leftmargin=*, noitemsep]
    \item \texthindi{उचित हिंदी शब्दावली का प्रयोग करें}
    \item \texthindi{संतुलित लंबाई के वाक्य बनाएं}
    \item \texthindi{विविध प्रकार के उदाहरण दें}
    \item \texthindi{विषय की गहरी समझ प्रदान करें}
    \item \texthindi{उत्तर विस्तृत और पूर्ण दें (6-8 वाक्य)}
    \item \texthindi{संदर्भ और व्याकरण की बारीकियों का उल्लेख करें}
\end{itemize}
\end{promptbox}

\paragraph{Level 3: Proficient}
\begin{promptbox}
You are a proficient Hindi language expert. Please answer in high-quality language:
\begin{itemize}[leftmargin=*, noitemsep]
    \item Use appropriate Hindi vocabulary.
    \item Make sentences of balanced length.
    \item Provide diverse examples.
    \item Provide a deep understanding of the topic.
    \item Give a detailed and complete answer (6-8 sentences).
    \item Mention context and grammatical nuances.
\end{itemize}
\end{promptbox}

\paragraph{Level 4: \texthindi{उन्नत} (Advanced)}
\begin{promptbox}
\texthindi{आप एक उच्च योग्यता प्राप्त हिंदी भाषा-शास्त्री हैं। उत्तर उन्नत स्तर की भाषा में दें:}
\begin{itemize}[leftmargin=*, noitemsep]
    \item \texthindi{परिष्कृत हिंदी शब्दावली का प्रयोग करें}
    \item \texthindi{जटिल व्याकरणिक संरचनाओं का उपयोग करें}
    \item \texthindi{अमूर्त और तकनीकी उदाहरण दें}
    \item \texthindi{गहन विश्लेषण प्रदान करें}
    \item \texthindi{उत्तर व्यापक और बहुआयामी दें (8-10 वाक्य)}
    \item \texthindi{भाषाविज्ञान के सिद्धांतों का प्रयोग करें}
    \item \texthindi{विषय के ऐतिहासिक और सांस्कृतिक संदर्भ दें}
\end{itemize}
\end{promptbox}

\paragraph{Level 4: Advanced}
\begin{promptbox}
You are a highly qualified Hindi linguist. Please answer in advanced-level language:
\begin{itemize}[leftmargin=*, noitemsep]
    \item Use sophisticated Hindi vocabulary.
    \item Use complex grammatical structures.
    \item Give abstract and technical examples.
    \item Provide in-depth analysis.
    \item Give a comprehensive and multi-dimensional answer (8-10 sentences).
    \item Use principles of linguistics.
    \item Provide historical and cultural context for the topic.
\end{itemize}
\end{promptbox}

\paragraph{Level 5: \texthindi{विशेषज्ञ} (Expert)}
\begin{promptbox}
\texthindi{आप एक अग्रणी हिंदी भाषाविज्ञान विशेषज्ञ हैं। उत्तर विशेषज्ञ स्तर की भाषा में दें:}
\begin{itemize}[leftmargin=*, noitemsep]
    \item \texthindi{उच्चतम स्तर की तकनीकी शब्दावली का प्रयोग करें}
    \item \texthindi{अत्यधिक जटिल भाषिक संरचनाओं का उपयोग करें}
    \item \texthindi{गहन तकनीकी और सैद्धांतिक उदाहरण दें}
    \item \texthindi{विषय का संपूर्ण और बहुस्तरीय विश्लेषण प्रदान करें}
    \item \texthindi{उत्तर अत्यंत विस्तृत और शोधपूर्ण दें (10+ वाक्य)}
    \item \texthindi{अत्याधुनिक भाषाविज्ञान अनुसंधान का संदर्भ दें}
    \item \texthindi{अंतःविषयक दृष्टिकोण अपनाएं}
    \item \texthindi{भाषा के दार्शनिक पहलुओं पर चर्चा करें}
\end{itemize}
\end{promptbox}

\paragraph{Level 5: Expert}
\begin{promptbox}
You are a leading Hindi linguistics expert. Please answer in expert-level language:
\begin{itemize}[leftmargin=*, noitemsep]
    \item Use the highest level of technical vocabulary.
    \item Use highly complex linguistic structures.
    \item Give deep technical and theoretical examples.
    \item Provide a complete and multi-layered analysis of the topic.
    \item Give an extremely detailed and research-oriented answer (10+ sentences).
    \item Reference cutting-edge linguistics research.
    \item Adopt an interdisciplinary approach.
    \item Discuss the philosophical aspects of language.
\end{itemize}
\end{promptbox}

\section{Dataset Generation Examples}
\label{sec:appendix_examples}

This appendix provides concrete examples of the instruction-response pairs generated by our pipeline, illustrating how structured WordNet data is converted into conversational formats.

\subsection{Basic Instructional Examples}
These examples focus on direct concept queries.
\begin{itemize}
    \item \textbf{Instruction:} \texthindi{"'प्रेम' का अर्थ क्या है?"} \textit{(What is the meaning of 'Love'?)}
    \item \textbf{Response:} \texthindi{"'प्रेम' का अर्थ है: गहरा स्नेह..."} \textit{(The meaning of 'Love' is: deep affection...)}
\end{itemize}

\subsection{Complex Multi-Aspect Examples}
These examples combine multiple semantic facets (definition, part-of-speech, synonyms, and specific relations) into a single, comprehensive response.
\begin{itemize}
    \item \textbf{Instruction:} \texthindi{"हिंदी शब्द 'सूर्य' के बारे में विस्तृत जानकारी दें।"} \textit{(Give detailed information about the Hindi word 'Sun'.)}
    \item \textbf{Response:} \texthindi{अर्थ: सौरमंडल का प्रधान ग्रह जो एक बहुत बड़ा और जलता हुआ गोला है।\\समानार्थी शब्द: सूरज, दिनकर, रवि, भानु\\उदाहरण: सूर्य पूर्व दिशा से निकलता है।\\व्याकरणिक श्रेणी: संज्ञा} \textit{(Meaning: The principal planet of the solar system... Synonyms: Suraj, Dinkar... Example: The sun rises in the east. POS: Noun)}

    \item \textbf{Instruction:} \texthindi{"हिंदी शब्द 'वृक्ष' के पर्यायवाची और अवयव संबंधों का वर्णन करें।"} \textit{(Describe the synonym and meronym relations for the Hindi word 'Tree'.)}
    \item \textbf{Response:} \texthindi{समानार्थी: पेड़, तरु, पादप, विटप\\अवयव\_सदस्य\_संग्रह: वन, जंगल\\अवयव\_घटक\_पदार्थ: जड़, तना, शाखा, पत्ता} \textit{(Synonyms: Ped, Taru... Member\_of\_collection: Forest... Component\_parts: Root, Trunk, Branch, Leaf)}
\end{itemize}

\subsection{Ontological Hierarchy Examples}
These examples teach the taxonomic structure of objects.
\begin{itemize}
    \item \textbf{Instruction:} \texthindi{"हिंदी शब्द 'गुलाब' के लिए वर्गीकरण पदानुक्रम क्या है?"} \textit{(What is the classification hierarchy for the Hindi word 'Rose'?)}
    \item \textbf{Response:} \texthindi{"'गुलाब' का वर्गीकरण: जीव -> पौधा -> फूल वाला पौधा -> गुलाब"} \textit{(Classification of 'Rose': Organism $\to$ Plant $\to$ Flowering Plant $\to$ Rose)}
\end{itemize}

\subsection{Disambiguation Examples}
These examples explicitly handle polysemy by presenting multiple distinct meanings for a single word form.
\begin{itemize}
    \item \textbf{Instruction:} \texthindi{"हिंदी शब्द 'आम' के विभिन्न अर्थों को समझाएं।"} \textit{(Explain the different meanings of the Hindi word 'Aam'.)}
    \item \textbf{Response:} \texthindi{'आम' के विभिन्न अर्थ हैं:\\\\अर्थ 1: एक प्रकार का प्रसिद्ध मीठा और रसीला फल\\व्याकरणिक श्रेणी: संज्ञा\\उदाहरण: तोता आम खा रहा है।\\\\अर्थ 2: जो विशिष्ट न होकर साधारण हो\\व्याकरणिक श्रेणी: विशेषण\\उदाहरण: यह आम रास्ता नहीं है।} \textit{(Meanings of 'Aam':\\Meaning 1: A famous sweet juicy fruit (Noun)...\\Meaning 2: Common or ordinary, not specific (Adjective)...)}
\end{itemize}

\end{document}